\documentclass[runningheads]{llncs}
\usepackage[T1]{fontenc}
\usepackage{graphicx}
\usepackage{booktabs}
\usepackage[misc]{ifsym}

\usepackage{mwe}
\usepackage[hyphens]{url}
\usepackage{caption}
\usepackage{amssymb}
\usepackage{amsmath}
\usepackage{svg}
\usepackage{subcaption}
\usepackage{algorithm}
\usepackage{algorithmic}
\usepackage{booktabs}
\usepackage{url}
\usepackage{xcolor}
\usepackage{longtable}
\usepackage{hyperref}
\usepackage[dvipsnames]{xcolor}
\begin{document}
\title{Towards Foundation Models for Consensus Rank Aggregation}
% \titlerunning{Foundation Model for Rank Aggregation}

\author{Yijun Jin\inst{1}\orcidID{0009-0000-1237-0648}\ \and Simon Klüttermann\inst{1}\orcidID{0000-0001-9698-4339} \and
Chiara Balestra\inst{1}\orcidID{0000-0002-3620-9012} \and
Emmanuel Müller\inst{1,2}\orcidID{0000-0002-5409-6875}}
\authorrunning{Yijun Jin et al.}
\institute{TU Dortmund University, Dortmund, Germany \and
Research Center Trustworthy Data Science and Security, University Alliance Ruhr, Dortmund, Germany\\
\email{yijun.jin@tu-dortmund.de (corresponding author)}\\
\email{Simon.Kluettermann@cs.tu-dortmund.de}}

\maketitle

\begin{abstract}
Aggregating a consensus ranking from multiple input rankings is a fundamental problem with applications in recommendation systems, search engines, job recruitment, and elections. Despite decades of research in consensus ranking aggregation, minimizing the Kemeny distance remains computationally intractable. Specifically, determining an optimal aggregation of rankings with respect to the Kemeny distance is an NP-hard problem, limiting its practical application to relatively small-scale instances. We propose the Kemeny Transformer, a novel Transformer-based algorithm trained via reinforcement learning to efficiently approximate the Kemeny optimal ranking. Experimental results demonstrate that our model outperforms classical majority-heuristic and Markov-chain approaches, achieving substantially faster inference than integer linear programming solvers. Our approach thus offers a practical, scalable alternative for real-world ranking-aggregation tasks.
\end{abstract}

\section{Introduction}

% TODO: Fill in introduction, motivation, and problem statement.
Consensus ranking aggregation is a fundamental problem concerned with merging multiple, often conflicting, rank-ordered lists into a single representative ranking that best reflects the collective preferences of the inputs. This challenge, with its origins in social choice theory and political elections~\cite{arrow1964social}, has gained significant traction across a diverse range of modern applications, including search engines~\cite{desarkar2016preference}, recommendation systems~\cite{oliveira2020rank}, biotechnology workflows~\cite{lin2010rank}, and machine learning~\cite{salman2022stability}.

A foundational principle in consensus ranking is the Condorcet criterion, which dictates that an item consistently winning pairwise comparisons should be ranked first~\cite{de1785essai}. Because cyclical preferences (the Condorcet paradox) can prevent a guaranteed winner, this concept is naturally extended to finding an optimal winning ranking~\cite{arrow1986social}. Among Condorcet-based methods, optimizing the Kemeny distance measured via pairwise Kendall-tau disagreements~\cite{kemeny1959mathematics,kemeny1962mathematical} stands out as the unique approach that simultaneously satisfies neutrality, consistency, and the Condorcet criterion~\cite{young1978consistent}, making the minimization of the total Kemeny distance the standard objective for ideal consensus aggregation.

Finding the Kemeny optimal ranking is known to be NP-hard, as established through a reduction to the Feedback Arc Set problem even in cases involving as few as four input permutations, thereby rendering exact algorithms computationally infeasible for large-scale instances~\cite{bartholdi1989voting,dwork2001rank}. As a result, numerous approximation methods have been proposed, including heuristic approaches~\cite{beck1983some}, pivot-based sorting algorithms~\cite{ailon2008aggregating}, Markov chain models~\cite{dwork2001rank}, and differential evolution approaches~\cite{d2017differential}. 

Intuitively, consensus ranking aggregation can be viewed as analogous to language translation tasks in natural language processing (NLP), given that both are fundamentally sequence-to-sequence tasks. However, they differ significantly in their evaluation metrics. NLP translation evaluation metrics are often ambiguous and subjectively defined, making precise optimization challenging. Conversely, ranking aggregation benefits from the clearly defined mathematical metric of the Kemeny distance. Unlike language translation, where massive historical human-generated data facilitates learning translation patterns, ranking aggregation lacks extensive datasets. Nevertheless, the explicit and mathematically well-defined nature of the Kemeny distance enables the development of self-teaching deep reinforcement learning (DRL) methods. Leveraging recent advances in DRL, such approaches can learn effective aggregation patterns to minimize the Kemeny distance without extensive pre-existing data.

The advent of the Transformer architecture in 2017 represents a seminal development in artificial intelligence~\cite{vaswani2017attention}. It has catalyzed a paradigm shift in natural language processing (NLP), elevating the performance of numerous sequence-to-sequence tasks to near-human proficiency across applications such as machine translation, text summarization, and question answering~\cite{kalyan2021ammus}. Central to the Transformer's success is its self-attention mechanism, which enables the model to dynamically weight the importance of different tokens within an input sequence, thereby capturing complex contextual relationships. This design principle gives Transformers an exceptional capacity to learn rich, context-aware mappings between sequences.

Consensus ranking is itself a structured sequence generation problem, producing a permutation that reflects complex, global preferences. Motivated by the Transformer’s demonstrated ability to model and generalize over sequence relationships efficiently, we introduce the {\textit{Kemeny Transformer}}: a sequence-to-sequence Transformer model featuring an auto-regressive decoder that incrementally generates a consensus ranking. The model is trained via reinforcement learning to directly minimize the expected Kemeny distance, thereby aligning the learning objective with the underlying aggregation criterion. Empirical evaluations demonstrate that our approach consistently outperforms a range of widely used approximation methods. Moreover, it achieves significant inference time speedups relative to exact integer linear programming (ILP) solvers~\cite{conitzer2006improved}, which, although capable of producing optimal Kemeny rankings, are hindered by exponential worst-case computational complexity. Our method also outperforms differential evolution algorithms~\cite{d2017differential}. These results highlight the scalability and practical utility of our approach for real-world consensus ranking aggregation.

Our main contributions can be summarized as follows:
\begin{itemize}
  \item We are the first to explore a sequence-to-sequence Transformer architecture for approximating the Kemeny optimal rank aggregation problem and discover a suitable reinforcement framework to train the model.
  \item We perform comprehensive empirical comparisons against classical approximation techniques, including heuristic procedures, pivot-based algorithms, Markov chain methods, and differential evolution algorithms, evaluating both ranking quality (Kemeny distance) and computational efficiency.
  \item We benchmark against integer linear programming solvers to quantify reductions in inference time and scalability gains, highlighting the practicality of our approach for large-scale ranking tasks.
  \item We demonstrate that our method generalizes well beyond the training distribution and can thus be used even for ranking problems it has never seen during training.
\end{itemize}

To increase reproducibility, our implementation, used datasets, and trained model checkpoints are available at \href{https://anonymous.4open.science/r/Kemeny-Transformer-ECBE/}{https://anonymous.4open.science/r/Kemeny-Transformer-ECBE/}.

\section{Related Work}

% TODO: Summarize existing ranking aggregation methods and transformer applications.

\subsection{Rank Aggregator}

Conitzer et al.~\cite{conitzer2006improved} formulated the Kemeny rank aggregation problem as an integer linear program (ILP), yielding exact, theoretically optimal solutions. However, this approach suffers from a super-polynomial worst-case runtime, rendering it impractical for large-scale instances.

To address scalability challenges, a variety of approximation methods have been proposed. Beck et al. introduced a greedy heuristic based on maximizing agreement and minimizing regret, which achieves a $2-3\%$ Kemeny distance gap to the optimal solution for ten-item instances~\cite{beck1983some}. However, its accuracy tends to deteriorate as the number of items increases.

Dwork et al.~\cite{dwork2001rank} proposed a Markov chain-based rank aggregation technique, which constructs a stationary distribution over items based on pairwise preferences. This method demonstrated superior empirical performance when compared to classical heuristics in applications such as web search result fusion and biomarker identification.

Ailon et al.~\cite{ailon2008aggregating} developed KiwiSort, a Quicksort-like algorithm that selects a pivot item and partitions the remaining items based on majority preferences in the base rankings: items that are preferred over the pivot are placed to the left, and others to the right. The process is applied recursively until a complete ranking is formed. KiwiSort is proven to be a 2-approximation algorithm with respect to the optimal Kemeny distance.

D'Ambrosio et al. proposed the Differential Evolution Consensus Ranking (DECoR) algorithm, which employs evolutionary strategies to identify rankings that minimize the Kemeny distance to a set of input base rankings~\cite{d2017differential}. The algorithm begins by initializing a population of item rankings and iteratively evolves this population using mutation and crossover operations, selecting individuals with lower Kemeny distances for survival. DECoR has been shown to outperform benchmark techniques such as simulated annealing in both solution quality and runtime on moderate-scale instances. However, its performance depends heavily on two parameters: the population size ($P$) and the allowed number of consecutive generations without improvement ($L$). Because achieving high-quality solutions typically requires large values for both, the computational cost increases substantially, limiting the algorithm's scalability for large-scale real-world applications.

\subsection{Transformers for Combinatorial Optimization}

In recent years, Transformer architectures~\cite{vaswani2017attention} and attention-based neural models~\cite{vinyals2015pointer} have demonstrated impressive performance on NP-hard combinatorial optimization problems. Kool et al.~\cite{kool2018attention} adapted a pointer network with attention mechanisms for routing problems by encoding node coordinates directly as input embeddings, omitting standard positional encoding, and employing an auto-regressive decoder to sequentially construct tours. Their approach achieved state-of-the-art results on the Traveling Salesman Problem (TSP), outperforming traditional heuristics and optimization techniques in both quality and speed.

Building on this foundation, Bresson et al.~\cite{bresson2021transformer} introduced a modified Transformer architecture specifically tailored to the TSP. Their model achieved an average optimality gap of just $1.42\%$ on instances with 100 nodes, with an average runtime of only 4.6 seconds. This represents a substantial efficiency gain over classical solvers, which often take several minutes to reach comparable solutions. These results underscore the potential of Transformer-based models as scalable, general-purpose solvers for complex combinatorial tasks.

\subsection{Positioning Our Approach}
Building on these successes, our work is the first to use Transformer models for consensus ranking. The Kemeny Transformer solves a major problem in past research: the need to choose between fast but inaccurate methods and accurate but slow methods like ILP or DECoR. By treating rank aggregation as a sequence-to-sequence task and training it with reinforcement learning, our model avoids the slow search processes of older algorithms. It uses an attention mechanism to understand the whole picture and generate highly accurate rankings in a single pass. As a result, our approach surpasses the speed limits of prior work and sets a new standard for fast, accurate, and practical consensus ranking.

\section{Methodology}
\begin{figure}[htb]
    \centering
    \includegraphics[width=0.9\textwidth]{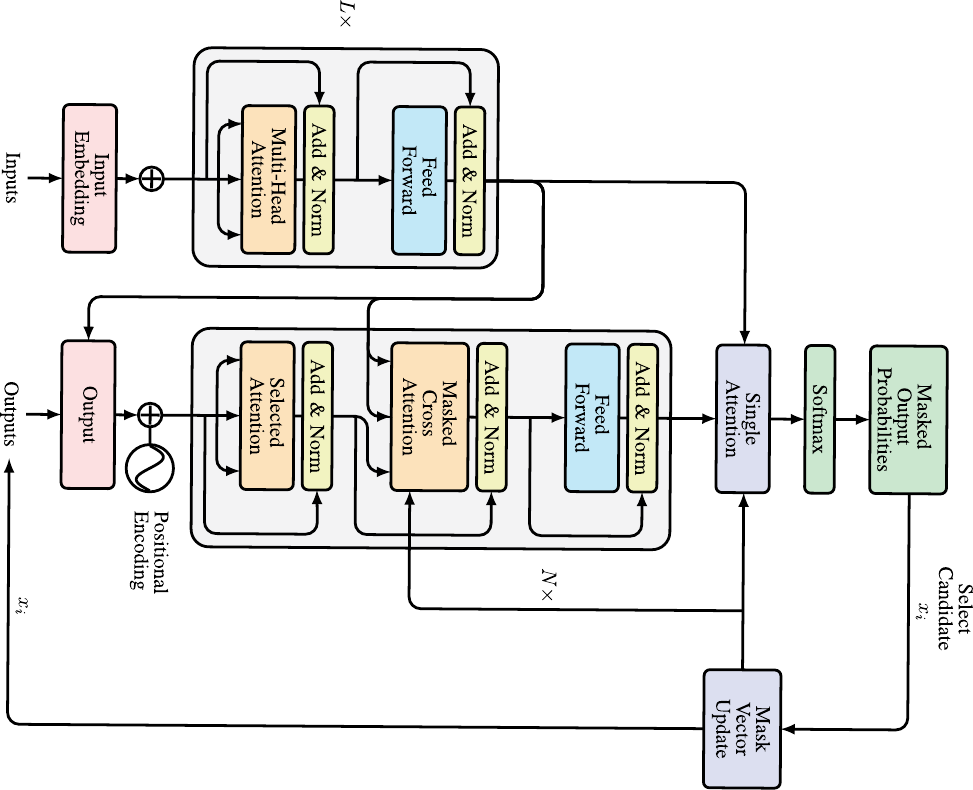}
    \caption{The Kemeny Transformer is organized as a classic encoder–decoder architecture. In this design, the encoder first processes the inputs to produce contextual representations, and the decoder then employs an auto-regressive procedure to sequentially generate a ranking that is close to Kemeny optimality.}
\end{figure}

\subsection{Problem Definition}
Let $X=\{x_1,\dots,x_n\}$ be a set of $n$ items. A ranking $\rho\in S_n$ is a permutation of these items, denoted as:
\[
  \rho = [x_{\rho(1)} \prec x_{\rho(2)} \prec \dots \prec x_{\rho(n)}],
\]
where $S_n$ is the set of all permutations over $X$. Given $m$ base rankings $R=\{\rho_1,\rho_2,\dots,\rho_m\} \subseteq S_n$, the goal is to derive a consensus ranking that best represents these inputs.

The Kendall-tau~\cite{kendall1938new} distance between two rankings $\rho$ and $\sigma$ counts the number of pairwise disagreements:
\begin{equation}
  K(\rho,\sigma)= \sum_{\{x_i,x_j\}\subseteq X}
    I\bigl((\rho(x_i)<\rho(x_j))\wedge(\sigma(x_j)<\sigma(x_i))\bigr),
\end{equation}
where $\rho(x_i)$ is the position of item $x_i$ in ranking $\rho$ and $I(\cdot)$ is the indicator function that equals 1 if an item pair $\{x_i,x_j\}$ is ordered differently in $\rho$ and $\sigma$, and 0 otherwise.

The Kemeny distance~\cite{kemeny1959mathematics} is defined as the average Kendall-tau distance from the ranking $\rho$ to the base rankings in $R$:
\begin{equation}
  d_K(\rho,R)= \frac{1}{|R|} \sum_{\sigma\in R} K(\rho,\sigma).
\end{equation}

The Kemeny optimal consensus ranking is the ranking $\rho^*\in S_n$ that minimizes the Kemeny distance:
\begin{equation}
  \rho^* = \arg\min_{\rho\in S_n} d_K(\rho,R)
\end{equation}

\subsection{Kemeny Transformer}

% TODO: Describe the proposed model architecture and innovations.
In this section, we will introduce the structure of our modified Transformer, denoted as the Kemeny Transformer. The structure is illustrated in Figure 1.

\subsection{Tokenization}
Unlike the vanilla Transformer, which utilizes one-hot encoding for token representation, our model treats each item $x_i$ as a distinct token. Specifically, each item is represented by a continuous feature vector $h_i^{\text{in}} \in \mathbb{R}^{m \times 1}$, where the elements correspond to the rank positions of item $x_i$ across the $m$ base rankings: $h_i^{\text{in}} = \bigl[p_1(x_i), p_2(x_i), \ldots, p_m(x_i)\bigr]$. To enhance the model's generalization capabilities across varying input dimensions, we normalize these positional vectors by dividing them by the maximum rank position (i.e., the total number of items, $n$). Because the fundamental objective of consensus aggregation relies on relative orderings rather than absolute positional values, this normalization preserves all essential ranking information without any loss of fidelity. Consequently, for an aggregation task comprising $n$ items, the complete input to the encoder is encoded in the matrix $H^{\text{in}} = [h_1^{\text{in}}, \ldots, h_n^{\text{in}}]^T \in \mathbb{R}^{n \times m}$.

\subsection{Encoder}
The encoder begins by projecting each input vector $h_i^{\text{in}}$ into a $d_{\text{model}}$-dimensional embedding vector via a linear layer. Positional embeddings are omitted, as the initial rank positions are already encoded in $H^{\text{in}}$. The resulting embeddings pass through $L$ layers of multi-head self-attention and feed-forward sub-layers, yielding contextualized representations $H^{\text{enc}}\in\mathbb{R}^{n\times d_{\text{model}}}$, which serve as hidden vectors for the decoder.

\subsection{Decoder}
Our auto-regressive decoder constructs the consensus ranking one item at a time. At decoding step $t$, let $x_{t-1}$ be the item selected at the previous step. Its encoder output $h_{t}^{\text{enc}}$ is retrieved and added to the positional embedding $PE(t,2d_i)$ to form the query embedding vector:
\begin{equation*}
     q_t = h_{i_t}^{\text{enc}} + PE(t,2d_i)
\end{equation*}
where $d_i$ is the dimension, and the query embedding vector is stored in the KV cache $H^{\text{sel}}=[q_1,q_2,...,q_t]$. This query attends over all previous decoder outputs (to enforce prefix consistency) using a self-attention mechanism. The output $h_{t}^{sel}$ is computed as follows:
 \begin{equation*}
     q_{t}^{\text{sel}}=h_{t}W_q^{\text{sel}}, \quad K_t^{\text{sel}}=H^{\text{sel}}W_K^{\text{sel}}, \quad  V_t^{\text{sel}}=H^{\text{sel}}W_V^{\text{sel}}
 \end{equation*}
 \begin{equation*}
     h_{t}^{\text{sel}}=softmax\left(\frac{q_{i_t}^{\text{sel}}{K_t^{\text{sel}}}^T}{\sqrt{d_{\text{model}}}}\right)V_t^{\text{sel}} 
 \end{equation*}

Moving to the subsequent layer, a masked cross-attention mechanism is employed to compute the attention of $h_{t}^{sel}$ over unselected items, facilitating the selection of the next prospective item from the unselected pool. The output $h_{t}^{cross}$ is determined by:
\begin{equation*}
     q_{t}^{\text{cross}}=h_{t}^{\text{sel}}W_q^{\text{unsel}}, \quad   K^{\text{unsel}}=H^{enc}W_K^{\text{unsel}} 
     , \quad V^{\text{unsel}}=H^{enc}W_V^{\text{unsel}} 
 \end{equation*}
 \begin{equation*}
     h_{i_t}^{\text{cross}}=softmax\left(\frac{q_{i_t}^{\text{cross}}{K^{\text{unsel}}}^T}{\sqrt{d_{\text{model}}}} \odot \mathcal{M}ask_t\right)V_t 
 \end{equation*} 

This is followed by a feed-forward network, as in the vanilla Transformer, to generate the final hidden vector $h_{t}^{f}$. After $N$ layers of the above attention structure, the hidden vector $h_{t}^{f}$ goes through a single-head attention over the unselected encoder representations as $K_{t}^f$ to produce logits. The output ${\text{logits}}_t$ is calculated as follows:
 \begin{equation*}
    {\text{logits}}_t=\frac{q_{t}^{f}{K_{t}^{f}}^T}{\sqrt{d_{\text{model}}}} \odot \mathcal{M}ask_t \in  \mathbb{R}^{n} 
 \end{equation*} 

After applying softmax on the logits, a categorical distribution over the remaining items is generated. The next item $x_{t}$ is sampled or chosen greedily from this distribution, and the model updates the mask vector by masking the selected item. The process then repeats until all $n$ positions are filled. 

\subsection{Training Method}
During training, we employ the REINFORCE algorithm~\cite{williams1992simple} with a self-critical roll-out baseline~\cite{rennie2017self}, as detailed in Algorithm 1. Both the training and baseline models are initialized randomly. At each training step, a batch of size $B$ of instances is sampled. For each instance $R_i$, the training model produces a ranking $\pi_i$ via SampleRollout (the next item is probabilistically sampled according to the model's predicted distribution), while the baseline model generates a ranking $\pi_i^{\mathrm{BL}}$ via GreedyRollout (the item with the highest probability is strictly chosen at each step). We then compute their respective Kemeny distances $d_K(\pi_i,R_i)$ and $d_K(\pi_i^{\mathrm{BL}},R_i)$. The loss is estimated by:

\begin{equation*}
    \nabla\mathcal{L}
        \;\leftarrow\;\sum_{i=1}^{B}\bigl(d_K(\pi_i,R_i)-d_K(\pi_i^{BL},R_i)\bigr)\,\nabla_{\theta}\log p_{\theta}(\pi_i)
\end{equation*}
The training model parameters are updated using the Adam optimizer.

After each epoch of training, the training model's performance is evaluated. Both models are evaluated on a validation dataset using greedy rollout. If the Kemeny distance of the ranking generated by the training model to the base rankings is significantly smaller than that of the baseline model, the baseline model is replaced by the current trained model.

\begin{algorithm}[ht]
\caption{REINFORCE with Rollout Baseline}
\label{alg:reinforce-rollout-baseline}
\begin{algorithmic}[1]
  \REQUIRE number of epochs $E$, steps per epoch $T$, batch size $B$, significance level $\alpha$, validation samples $R^{v}$
  \ENSURE updated policy parameters $\theta$
  \STATE Initialize $\theta$ and baseline parameters $\theta^{BL}$
  \FOR{$epoch \leftarrow 1$ \TO $E$}
    \FOR{$step \leftarrow 1$ \TO $T$}
      \STATE $ R_i\leftarrow \text{RandomInstance}(),\;\forall i\in\{1,\dots,B\}$
      \STATE $\pi_i\leftarrow \text{SampleRollout}(R_i, p_{\theta}),\;\forall i\in\{1,\dots,B\}$
      \STATE $\pi_i^{B}\leftarrow \text{GreedyRollout}(R_i, p_{\theta^{B}}),\;\forall i\in\{1,\dots,B\}$
      \STATE Compute Kemeny distance gap:
      \STATE $\Delta_{i} \leftarrow d_K(\pi_i,R_i)-d_K(\pi_i^{BL},R_i) $
      \STATE Compute loss:
      \STATE $\nabla\mathcal{L}\leftarrow\sum_{i=1}^{B}\bigl(\Delta_{i}\bigr)\,\nabla_{\theta}\log p_{\theta}(\pi_i)$
      \STATE Update policy: $\theta \leftarrow \text{Adam}(\theta, \nabla\mathcal{L})$
    \ENDFOR
    \STATE $\text{significance} \leftarrow \text{OneSidedPairedTTest}(p_{\theta},p_{\theta^{BL}})  $
    \IF{$\text{significance} <\alpha$}
      \STATE Update baseline: $\theta^{BL} \leftarrow \theta$
    \ENDIF
  \ENDFOR
\end{algorithmic}
\end{algorithm}

\section{Experiments}
% TODO: Detail datasets, baselines, evaluation metrics, and results.

\subsection{Hyperparameters}

The Kemeny Transformer used in our evaluations consists of three encoder layers and two decoder layers. Input embeddings have a dimensionality of \(d_{\mathrm{model}}=128\), and each feed-forward sub-layer has an intermediate dimension of 512. The multi-head attention modules employ eight attention heads.

The model is trained for 250 epochs, each comprising 500 steps with a batch size of 
\(B=1024\). We use the Adam optimizer with a fixed learning rate of \(\eta=10^{-4}\). Training is conducted on two NVIDIA RTX 4500 Ada GPUs, requiring approximately 14 minutes per epoch.

The DECoR baseline is implemented using the R package ConsRank, with default hyperparameters: population size \(P=15\) and a stopping criterion based on \(L=100\) consecutive generations without improvement.

All methods are executed on an AMD Ryzen Threadripper PRO 5975WX with 32 physical cores and 512 GB of RAM. For all experiments, the random seed is fixed to 1234.

\subsection{Datasets}
We evaluate our model on three types of synthetically generated base rankings, each comprising eight permutations over $n=100$ items:
\begin{enumerate}
  \item \textbf{Random base rankings:} Each of the eight rankings is generated by independently sampling a uniform random permutation. This setting yields minimal inherent consensus, resulting in larger optimal Kemeny distances.
\item \textbf{Repeat base rankings:} A subset of the eight rankings is an exact duplicate of a reference permutation, while the remainder are independently sampled. This scenario captures varying levels of consensus; when all eight rankings coincide, the Kemeny distance of the consensus ranking is zero.

\item \textbf{Jiggling base rankings:} These datasets simulate scenarios with a strong underlying consensus by applying small, random local perturbations (or ``jiggling'') to a single reference permutation, \(\rho_{\mathrm{ref}}\). To generate each new base ranking, we let \(p_i\) denote the original position of an item \(x_i\) in \(\rho_{\mathrm{ref}}\). We then probabilistically select a target position \(p_j\) to swap it with. To ensure items mostly stay near their original spots, the probability of choosing target position \(p_j\) is exponentially penalized by the distance between the two positions:
\begin{equation*}
P(j \mid x_i)
= \frac{\exp\!\left(M - |p_j - p_i|\right)}
       {\sum_{p_k \neq p_i} \exp\!\left(M - |p_k - p_i|\right)},
\end{equation*}
where \(M\) is a constant scaling factor. Because the distance \(|p_j - p_i|\) is subtracted in the exponent, nearby positions have a significantly higher probability of being chosen for a swap. This procedure guarantees that items only shift slightly from their reference positions, producing a set of highly concordant base rankings with relatively small Kemeny distances between them.

  \item \textbf{Real-world ranking:} We compiled country and region rankings based on eight key metrics from the year 2020: Population, Human Development Index (HDI), Global Peace Index (GPI), Corruption Perceptions Index (CPI), Global Innovation Index (GII), Democracy Index, Gross Domestic Product (GDP), and World Trade Value. The data were sourced from the United Nations Data Portal~\cite{unpopulation2020} and Kaggle~\cite{kaggle}. We selected the top 100 most populous countries as our primary set, excluding certain countries or regions due to data unavailability or political instability. This dataset serves as the real-world benchmark for evaluating the performance of different ranking aggregation methods.
\end{enumerate}

\subsection{Evaluation}

We evaluate the Kemeny Transformer on a comprehensive test suite comprising 2,000 random rankings, 10,000 repeat rankings, and 10,000 jiggling rankings.

Performance is assessed using two principal metrics: the gap in Kemeny distance relative to the optimal solution produced by an integer linear programming solver, and the inference runtime.

To contextualize the effectiveness of our approach, we compare the Kemeny Transformer against established baseline methods, including the exact integer programming solver (Gurobi) and other widely used approximate aggregation techniques. The results highlight the Kemeny Transformer's ability to deliver high-quality consensus rankings with substantially improved computational efficiency, making it a scalable alternative for large-scale applications.

\begin{table*}[ht]
\begin{tabular}{l| ccc| ccc| ccc|}
\toprule
 & \multicolumn{3}{c}{random 2k} & \multicolumn{3}{c}{repeat 10k} & \multicolumn{3}{c}{jiggling 10k} \\

Method     & Obj   & Gap  & Time  & Obj   & Gap  & Time & Obj   & Gap  & Time\\ 
\midrule
Gurobi (exact)       &1858.48 &0 &25.12h  & 1245.01 & 0 &46.14h &731.79 &0 & 23.71h  \\\midrule

Kiwisort       &1980.13 &121.65 & 1.24m  & 1343.64 & 98.63 &6.15m &785.61 &53.82 & 6.14m  \\

Markov Chain &1894.02 &35.54 &20s  &1349.42 &104.41 &89.65s  &823.15 & 91.36 & 90.23s    \\

Max Agreement &1884.92 &26.44 &1.23m &1314.77 &69.75 &6.27m &788.76 &56.97 &6.28m\\

Min Regret &1885.17 &26.68 &1.31m &1309.39 &64.37 &6.42m &783.96 &52.17 &6.61m\\

DECoR &1891.65 &33.17 &94.47m &1282.89 &37.88 &9.33h &757.69 &25.90 &8.75h\\

Transformer(Ours)
&1873.20 &{\textbf{17.71}} &\textbf{0.88s} &1265.44 &{\textbf{20.43}} &\textbf{1.92s}&744.45 &{\textbf{12.66}} &\textbf{1.92s}   \\
\bottomrule
\end{tabular}
\caption{Kemeny Transformer vs. baselines. ``Obj'' denotes the average Kemeny distance of the rankings produced by each method. ``Gap'' measures the absolute and relative difference of each method’s Kemeny distance from the optimal solution obtained via Gurobi. ``Time'' indicates the total runtime each method requires to process the evaluation dataset. \textbf{The Kemeny Transformer is both more accurate and faster than each competing approximation method.}}
\label{tab:1}
\end{table*}
\section{Results}
\subsection{Results on Synthetic Datasets}

\begin{figure*}[ht]
  \centering
  \begin{subfigure}[b]{0.3\textwidth}
    \centering
    \includegraphics[width=\textwidth]{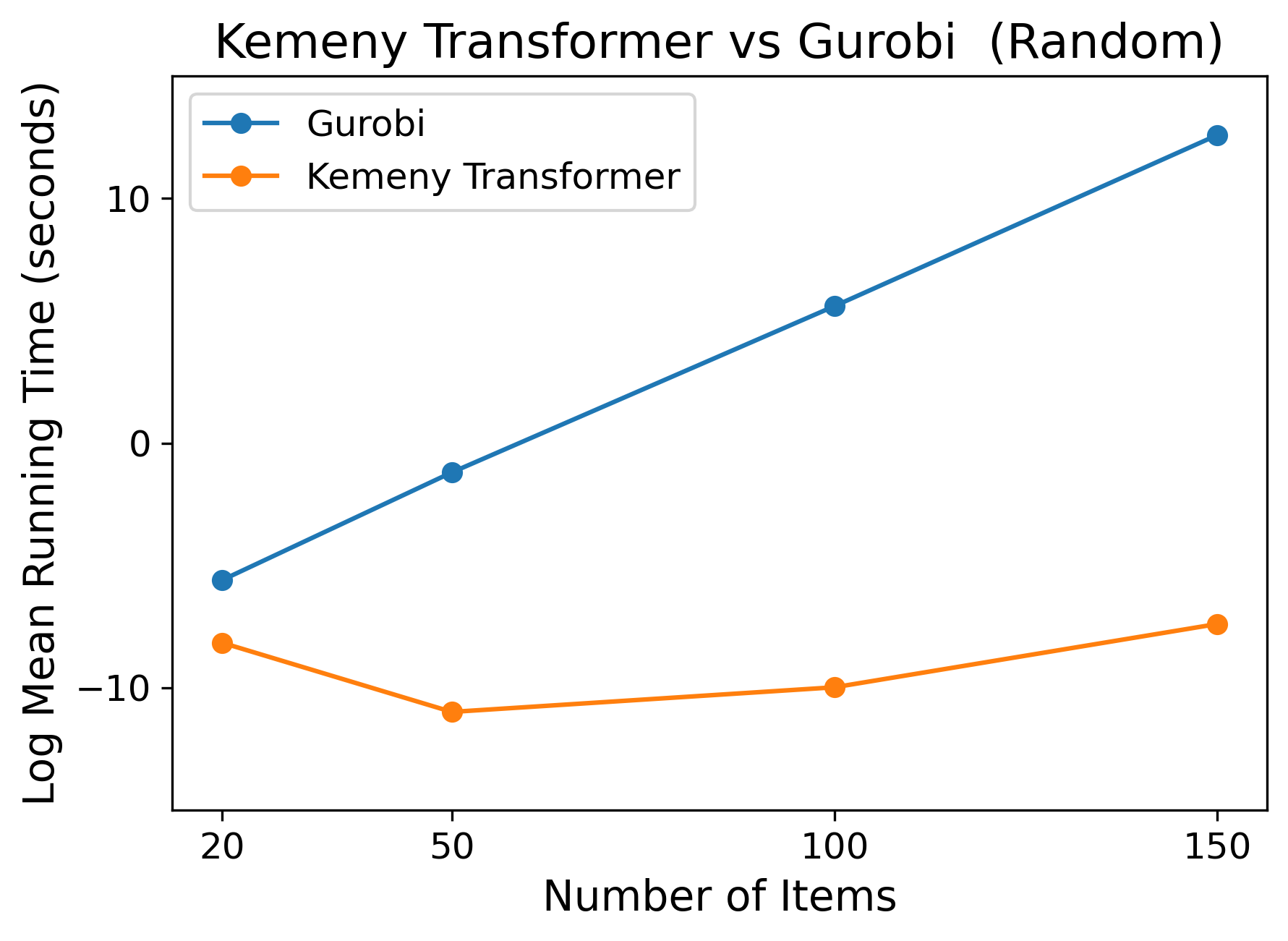}
    \caption{Running time on random datasets}
    \label{fig:sub1}
  \end{subfigure}
  \begin{subfigure}[b]{0.3\textwidth}
    \centering
    \includegraphics[width=\textwidth]{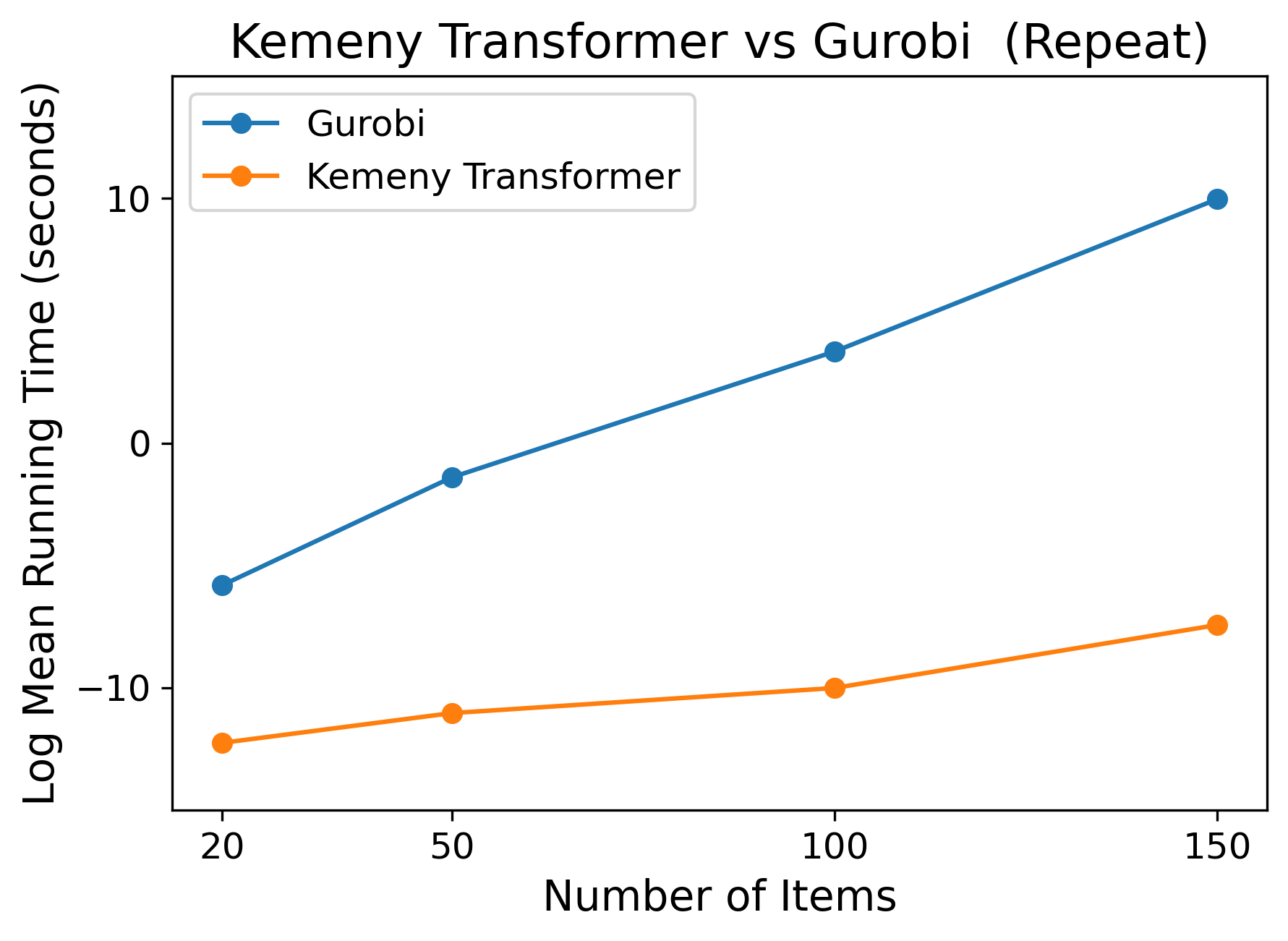}
    \caption{Running time on repeat datasets}
    \label{fig:sub2}
  \end{subfigure}
  \begin{subfigure}[b]{0.3\textwidth}
    \centering
    \includegraphics[width=\textwidth]{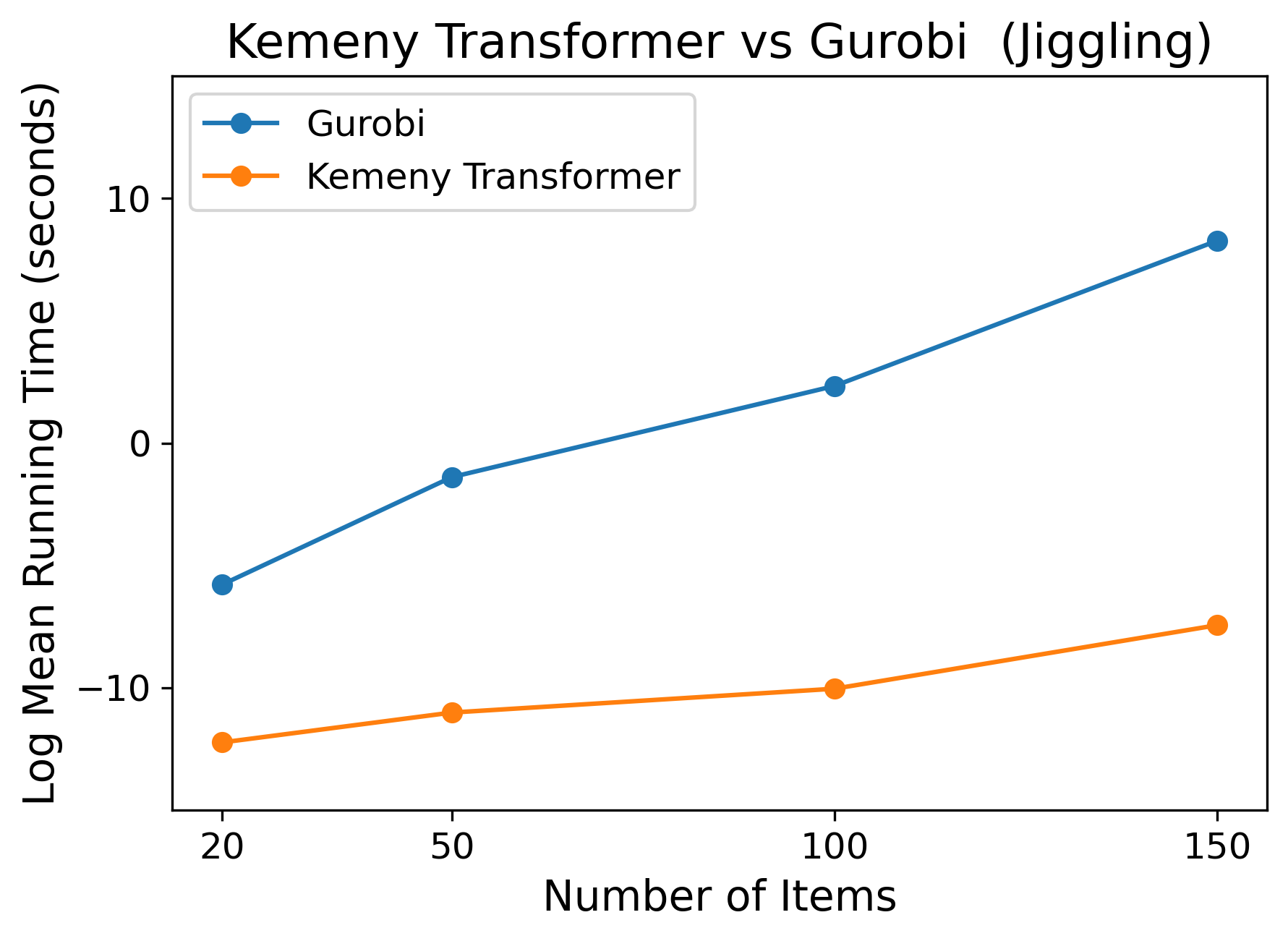}
    \caption{Running time on jiggling datasets}
    \label{fig:sub3}
  \end{subfigure}
  \caption{Running time comparison between the Kemeny Transformer and the Gurobi solver as the length of the base rankings increases. The x-axis denotes the number of items in the base rankings; the y-axis shows the logarithm of the mean running time computed over multiple trials. \textbf{Our model scales more efficiently to larger ranking problems.}}
  \label{scalability}
\end{figure*}
Table 1 presents a comprehensive summary of the experimental results obtained from evaluating our model on a suite of large-scale synthetic datasets. The results computed by Gurobi are treated as the ground-truth optimal Kemeny rankings, and accordingly, the Kemeny distance gap for Gurobi is set to zero.

The Kemeny Transformer consistently achieves lower mean Kemeny distance gaps compared to other approximate methods, demonstrating its superior ability to approximate the optimal consensus ranking. Moreover, it exhibits substantially reduced runtime across all baseline methods. This efficiency advantage can be attributed to the inherent design of the Transformer architecture, which is well-suited to parallel execution on GPUs, unlike many traditional heuristic and evolutionary algorithms that lack optimized parallel implementations.

In summary, the experimental results confirm the effectiveness of the Kemeny Transformer in producing high-quality consensus rankings with strong trade-offs between optimality and computational efficiency. The model demonstrates robust performance across diverse synthetic datasets, highlighting its adaptability and practical utility for large-scale consensus ranking tasks.

\begin{table*}[ht]
\centering
\fontsize{10}{12}\selectfont
\setlength{\tabcolsep}{1mm}
\begin{tabular}{|p{1.9cm}|p{1.9cm}|p{1.9cm}|p{1.9cm}|p{1.9cm}|p{1.9cm}|}
\hline
\textbf{Gurobi} & \textbf{Kemeny Transformer (Our)} & \textbf{Markov Chain} & \textbf{Kiwisort} & \textbf{Minimal Regret} & \textbf{DECoR} \\
\hline
\multicolumn{6}{|c|}{\textbf{Kemeny distance}} \\
\hline
997.63 & 1006.88 & 1039.63 & 1027.38 & 1031.38 & 1015.13 \\
\hline
\colorbox{Aquamarine}{\textbf{Swiss}} & \colorbox{Aquamarine}{\textbf{Germany}} & \colorbox{Aquamarine}{\textbf{Canada}} & \colorbox{Aquamarine}{\textbf{Germany}} & \colorbox{Aquamarine}{\textbf{Canada}} & \colorbox{Aquamarine}{\textbf{Germany}} \\
\colorbox{Aquamarine}{\textbf{Germany}}  & \colorbox{Aquamarine}{\textbf{Canada}} & \colorbox{Aquamarine}{\textbf{Germany}} & \colorbox{Aquamarine}{\textbf{Canada}} & Australia & \colorbox{Aquamarine}{\textbf{Swiss}} \\
\colorbox{Aquamarine}{\textbf{Canada}}  & \colorbox{Aquamarine}{\textbf{Swiss}} & Australia & \colorbox{Aquamarine}{\textbf{Swiss}} & \colorbox{Aquamarine}{\textbf{Germany}} & Netherlands \\
Sweden  & UK & USA& Australia & UK & UK \\
Netherlands & USA& UK & Sweden & USA& Sweden \\
Australia  & Australia & Sweden & Netherlands & Japan & USA\\
UK & Sweden & Japan & UK & France & \colorbox{Aquamarine}{\textbf{Canada}} \\
USA& Netherlands & \colorbox{Aquamarine}{\textbf{Swiss}} & \colorbox{red}{\textbf{S. Korea}} & Sweden & Japan \\
Japan  & Japan & France & Austria &\colorbox{red}{\textbf{Italy}} & France \\
France  & France & Netherlands & USA& \colorbox{Aquamarine}{\textbf{Swiss}} & \colorbox{red}{\textbf{S. Korea}} \\
\hline
\end{tabular}
\caption{Top 10 Country Rankings by Various Methods with Kemeny distances}
\label{tab:2}
\end{table*}

\subsection{Results on Real-World Data}
Table 2 presents the results of each method evaluated on a real-world dataset comprising the top 100 countries and regions ranked according to eight key metrics. In addition to reporting the Kemeny distance and its gap from the optimum, the table includes the top 10 ranked countries from each method’s output. Due to space constraints, the full rankings of all 100 countries are provided in the supplementary material. Including the top-ranked subset offers a more intuitive and interpretable view of each method's ranking behavior.

Consistent with earlier results, the Kemeny Transformer attains the smallest gap in Kemeny distance, outperforming all other approximate methods in terms of alignment with the optimal ranking obtained via Gurobi. A detailed inspection shows that Switzerland, Germany, and Canada occupy the top three positions in the optimal ranking, and the Transformer likewise ranks them within its top three. In contrast, the Markov Chain, Minimal Regret, and DECoR methods each omit at least one of these in their highest tiers, relegating it to a lower position and thereby contributing to an increased Kemeny distance. Furthermore, countries such as South Korea and Italy, absent from the top ten in the optimal ranking, appear within the top ten of the rankings produced by Kiwisort, DECoR, and Minimal Regret, which also inflates their respective Kemeny distances. The Kemeny Transformer avoids both of these unfavorable ranking patterns.

These findings indicate that the Kemeny Transformer not only performs robustly on synthetic benchmarks but also generalizes effectively to complex, real-world ranking aggregation tasks.

\subsection{Scalability}
To evaluate the scalability of the proposed method, we conducted a comparative analysis of inference times. The evaluation was performed on 100 instances for each of the synthetic datasets (random, jiggling, and repeat) with the number of items, \(n\), varying across the set {20, 50, 100, 150}.

As depicted in Figure 2, the inference time of the Kemeny Transformer demonstrates a marked advantage over Gurobi as the number of items increases. The runtime of the Kemeny Transformer grows significantly slower than that of the exact ILP solver. This empirical observation is consistent with theory: the Kemeny Transformer runs in $O(d_{\text{model}} n^2)$, i.e., quadratic in the number of items for a fixed model dimension. In contrast, Gurobi, which utilizes a branch-and-bound algorithm, has a complexity in the order of \(O(n^{2.5}b_n)\)~\cite{morrison2016branch}, where \(b_n\) represents the number of branches explored. This term can grow exponentially in the worst-case scenario.

These results highlight the superior scalability of the Kemeny Transformer. Its ability to efficiently handle large-scale instances makes it a practical solution for consensus rank aggregation in contexts where the computational cost of exact solvers becomes prohibitive.

\section{Generalization beyond Input Constraints}
In this section, we investigate the robustness and generalization capabilities of the Kemeny Transformer across varying numbers of voters and base ranking lengths. Specifically, we train a unified Kemeny Transformer model using mixed-batch training, meaning each training batch contains diverse instances with voter counts ranging from $6$ to $10$ and item counts spanning from $90$ to $110$. To rigorously evaluate its adaptability, we test this single trained model across nine distinct dataset configurations, combining voter counts $m \in \{6, 8, 10\}$ and item counts $n \in \{100, 125, 150\}$. While the tests with a number of items of 100 are within our training scope, the item counts of 125 and 150 are outside our training input size range.% By comparing the resulting Kemeny distances against traditional methods, we highlight the remarkable potential of the Kemeny Transformer to handle dynamic and diverse input sizes seamlessly.
%I remove this line, because it already gives the results before we show them

\subsection{Evaluation}
\begin{figure*}[ht]
  \centering
  
    \centering
    \includegraphics[width=\textwidth]{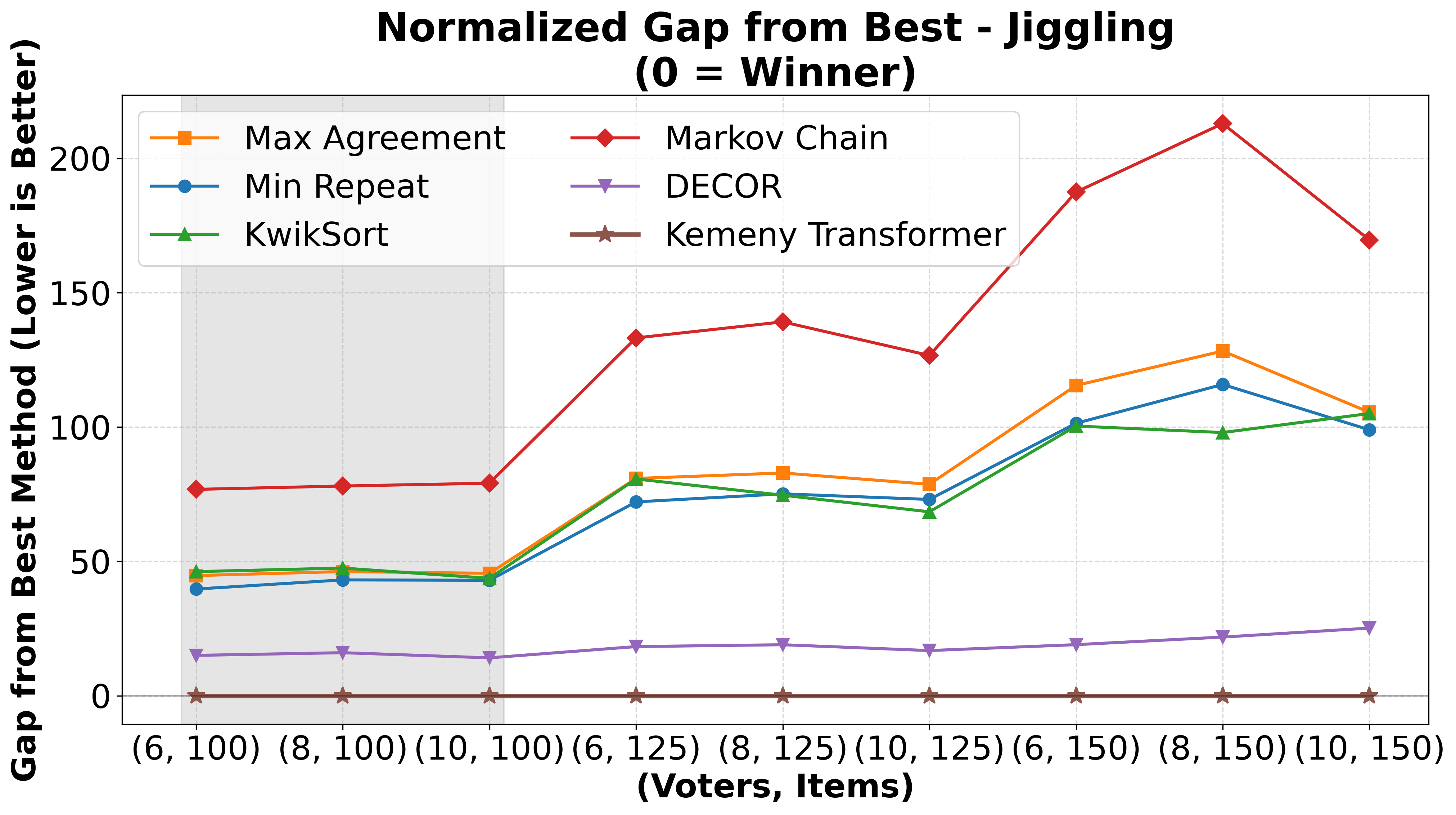}
  \caption{Performance comparison across varying input dimensions (voters, items) for the Jiggling datatype. The remaining datatypes are shown in the supplementary material. To ensure a fair comparison across different scales, the y-axis shows the normalized Kemeny distance gap, with the best-performing method serving as the baseline ($0$). Settings within the gray region are included in our training scope, while the remaining settings are outside it. \textbf{Both inside and outside of our training prior, the Kemeny Transformer achieves state-of-the-art performance.}. A possible reason for this good generalization stems from the input normalization used.}
  \label{fig:ablation}
\end{figure*}
For a clearer and more intuitive visualization, we evaluate the methods in Figure~\ref{fig:ablation} using a normalized Kemeny distance gap rather than the absolute Kemeny distance, which naturally scales with the problem size. The best-performing method for any given size configuration is established as the baseline (a gap of zero), and the performance of all other models is measured as the relative deviation from this optimal score\footnote{Because of time constraints, we are unable to calculate exact Kemeny optimal rankings here.}. The settings in the gray area are part of our training input size distribution, where our model performs best, as shown before. However, the performance of our model outside the training distribution is also better than that of each competitor and maintains a stable lead even with $50\%$ more items. Thus, our method can be used well beyond the training distribution, paving the way for consensus-aggregation foundation models.

% As strikingly illustrated in Figure \ref{fig:ablation}, the Kemeny Transformer achieves state-of-the-art performance across all tested dataset distributions: random, repeat, and jiggling. In every single evaluation scenario, our model sets the optimal baseline, maintaining a normalized gap of exactly zero regardless of the varying voter and item counts. This phenomenal consistency starkly contrasts with traditional algorithms, which exhibit significant and erratic performance degradation as input dimensions shift. These results definitively prove that the Kemeny Transformer is not only highly accurate but possesses exceptional robustness and immense potential to scale, cementing its position as a powerfully versatile solution for dynamic rank aggregation tasks.

\section{Conclusion}

In this work, we introduced the \textit{Kemeny Transformer}, representing a pivotal step toward establishing a foundation model paradigm for consensus rank aggregation under the Kemeny distance criterion. By framing rank aggregation as a sequence-to-sequence task trained via deep reinforcement learning, our model fundamentally departs from the limitations of traditional methods. While classical heuristics, evolutionary algorithms like DECoR, and exact integer linear programming solvers struggle with severe trade-offs between computational efficiency and solution optimality, our architecture leverages the global context-capturing capabilities of self-attention to decisively overcome these barriers. Through extensive evaluations on both synthetic and real-world datasets, the Kemeny Transformer consistently generated near-optimal rankings, outperforming traditional baselines in both accuracy and inference speed. Furthermore, our ablation studies demonstrated the model's remarkable robustness and zero-shot scalability across varying voter and item dimensions. These results strongly establish the Kemeny Transformer not just as a specialized tool, but also as a generalizable, highly scalable foundation model capable of dominating complex, large-scale consensus rank-aggregation tasks.

\begin{credits}
\subsubsection{\ackname}
This research was in part funded by the Research Center Trustworthy Data Science and Security (\url{https://rc-trust.ai}), one of the Research Alliance centres within the University Alliance Ruhr (\url{https://uaruhr.de}).

The research was further supported with computing time provided on LiDO3 (\url{https://lido.itmc.tu-dortmund.de/lido3/}).

\end{credits}

\bibliographystyle{splncs04}
\bibliography{Reference}

\begin{thebibliography}{10}
\providecommand{\url}[1]{\texttt{#1}}
\providecommand{\urlprefix}{URL }
\providecommand{\doi}[1]{https://doi.org/#1}

\bibitem{ailon2008aggregating}
Ailon, N., Charikar, M., Newman, A.: Aggregating inconsistent information: ranking and clustering. Journal of the ACM (JACM)  \textbf{55}(5),  1--27 (2008)

\bibitem{arrow1986social}
Arrow, K.J., Raynaud, H.: Social choice and multicriterion decision-making. MIT Press Books  \textbf{1} (1986)

\bibitem{arrow1964social}
Arrow, K.J., et~al.: Social choice and individual values, vol.~2. Wiley New York (1964)

\bibitem{bartholdi1989voting}
Bartholdi, J., Tovey, C.A., Trick, M.A.: Voting schemes for which it can be difficult to tell who won the election. Social Choice and welfare  \textbf{6},  157--165 (1989)

\bibitem{beck1983some}
Beck, M.P., Lin, B.W.: Some heuristics for the consensus ranking problem. Computers \& Operations Research  \textbf{10}(1), ~1--7 (1983)

\bibitem{bresson2021transformer}
Bresson, X., Laurent, T.: The transformer network for the traveling salesman problem. arXiv preprint arXiv:2103.03012  (2021)

\bibitem{de1785essai}
de~Caritat~Mis, J.A.N., et~al.: Essai sur l'application de l'analyse {\`a} la probabilit{\'e} des d{\'e}cisions rendues {\`a} la pluralit{\'e} des voix. Imprimerie royale (1785)

\bibitem{conitzer2006improved}
Conitzer, V., Davenport, A., Kalagnanam, J.: Improved bounds for computing kemeny rankings. In: AAAI. vol.~6, pp. 620--626 (2006)

\bibitem{desarkar2016preference}
Desarkar, M.S., Sarkar, S., Mitra, P.: Preference relations based unsupervised rank aggregation for metasearch. Expert Systems with Applications  \textbf{49},  86--98 (2016)

\bibitem{dwork2001rank}
Dwork, C., Kumar, R., Naor, M., Sivakumar, D.: Rank aggregation methods for the web. In: Proceedings of the 10th international conference on World Wide Web. pp. 613--622 (2001)

\bibitem{d2017differential}
D’Ambrosio, A., Mazzeo, G., Iorio, C., Siciliano, R.: A differential evolution algorithm for finding the median ranking under the kemeny axiomatic approach. Computers \& Operations Research  \textbf{82},  126--138 (2017)

\bibitem{kaggle}
{Kaggle}: Kaggle: Your machine learning and data science community (2020), \url{https://www.kaggle.com/}, accessed: 2023-10-25

\bibitem{kalyan2021ammus}
Kalyan, K.S., Rajasekharan, A., Sangeetha, S.: Ammus: A survey of transformer-based pretrained models in natural language processing. arXiv preprint arXiv:2108.05542  (2021)

\bibitem{kemeny1959mathematics}
Kemeny, J.G.: Mathematics without numbers. Daedalus  \textbf{88}(4),  577--591 (1959)

\bibitem{kemeny1962mathematical}
Kemeny, J.G.: Mathematical models in the social sciences  (1962)

\bibitem{kendall1938new}
Kendall, M.G.: A new measure of rank correlation. Biometrika  \textbf{30}(1-2),  81--93 (1938)

\bibitem{kool2018attention}
Kool, W., Van~Hoof, H., Welling, M.: Attention, learn to solve routing problems! arXiv preprint arXiv:1803.08475  (2018)

\bibitem{lin2010rank}
Lin, S.: Rank aggregation methods. Wiley Interdisciplinary Reviews: Computational Statistics  \textbf{2}(5),  555--570 (2010)

\bibitem{morrison2016branch}
Morrison, D.R., Jacobson, S.H., Sauppe, J.J., Sewell, E.C.: Branch-and-bound algorithms: A survey of recent advances in searching, branching, and pruning. Discrete Optimization  \textbf{19},  79--102 (2016)

\bibitem{oliveira2020rank}
Oliveira, S.E., Diniz, V., Lacerda, A., Merschmanm, L., Pappa, G.L.: Is rank aggregation effective in recommender systems? an experimental analysis. ACM Transactions on Intelligent Systems and Technology (TIST)  \textbf{11}(2),  1--26 (2020)

\bibitem{rennie2017self}
Rennie, S.J., Marcheret, E., Mroueh, Y., Ross, J., Goel, V.: Self-critical sequence training for image captioning. In: Proceedings of the IEEE conference on computer vision and pattern recognition. pp. 7008--7024 (2017)

\bibitem{salman2022stability}
Salman, R., Alzaatreh, A., Sulieman, H.: The stability of different aggregation techniques in ensemble feature selection. Journal of Big Data  \textbf{9}(1), ~51 (2022)

\bibitem{unpopulation2020}
{United Nations}: United nations data portal (2020), \url{https://population.un.org/dataportal/home?df=16a5d2ec-f118-4a0a-9cc7-7879c7925c6d}, accessed: 2023-10-25

\bibitem{vaswani2017attention}
Vaswani, A., Shazeer, N., Parmar, N., Uszkoreit, J., Jones, L., Gomez, A.N., Kaiser, {\L}., Polosukhin, I.: Attention is all you need. Advances in neural information processing systems  \textbf{30} (2017)

\bibitem{vinyals2015pointer}
Vinyals, O., Fortunato, M., Jaitly, N.: Pointer networks. Advances in neural information processing systems  \textbf{28} (2015)

\bibitem{williams1992simple}
Williams, R.J.: Simple statistical gradient-following algorithms for connectionist reinforcement learning. Machine learning  \textbf{8},  229--256 (1992)

\bibitem{young1978consistent}
Young, H.P., Levenglick, A.: A consistent extension of condorcet’s election principle. SIAM Journal on applied Mathematics  \textbf{35}(2),  285--300 (1978)

\end{thebibliography}

\end{document}